\documentclass[letterpaper, 10 pt, conference]{ieeeconf}  

\IEEEoverridecommandlockouts                              

\usepackage{hyperref} 
\usepackage{graphics} 
\usepackage{epsfig} 
\usepackage{times} 
\usepackage{amsmath} 
\usepackage{amssymb}  
\usepackage{cite}

\usepackage{booktabs}  %
\usepackage{colortbl}  %

\usepackage{subcaption}

\usepackage{tabulary}
\usepackage{multirow}

\begin{document}
\title{\LARGE \bf
Marrying NeRF with Feature Matching for One-step Pose Estimation
}
\author{
	  \begin{tabular}{ccc}
	Ronghan Chen\textsuperscript{1,2,3}\qquad & Yang Cong\textsuperscript{4}$^{\ast}$
	\qquad & Yu Ren\textsuperscript{1,2,3}
	  \end{tabular}
\\
\textsuperscript{1}State Key Laboratory of Robotics, Shenyang Institute of Automation, Chinese Academy of Sciences\thanks{This work is supported in part by the National Key Research and Development Program of China under Grant 2019YFB1310300 and the National Nature Science Foundation of China under Grant 61821005.}
\\
\textsuperscript{2}Institutes for Robotics and Intelligent Manufacturing, Chinese Academy of Sciences\\
\textsuperscript{3}University of Chinese Academy of Sciences\\
\textsuperscript{4}College of Automation Science and Engineering, South China University of Technology\\
{\tt\small chenronghan@sia.cn, congyang81@gmail.com, renyu0414@gmail.com}
}
\twocolumn[{
	\renewcommand\twocolumn[1][]{#1}%
	\maketitle
	\thispagestyle{empty}
	\pagestyle{empty}
	\begin{center}
		\centering
		\vspace{-15pt}
		\includegraphics[width=0.96\textwidth]{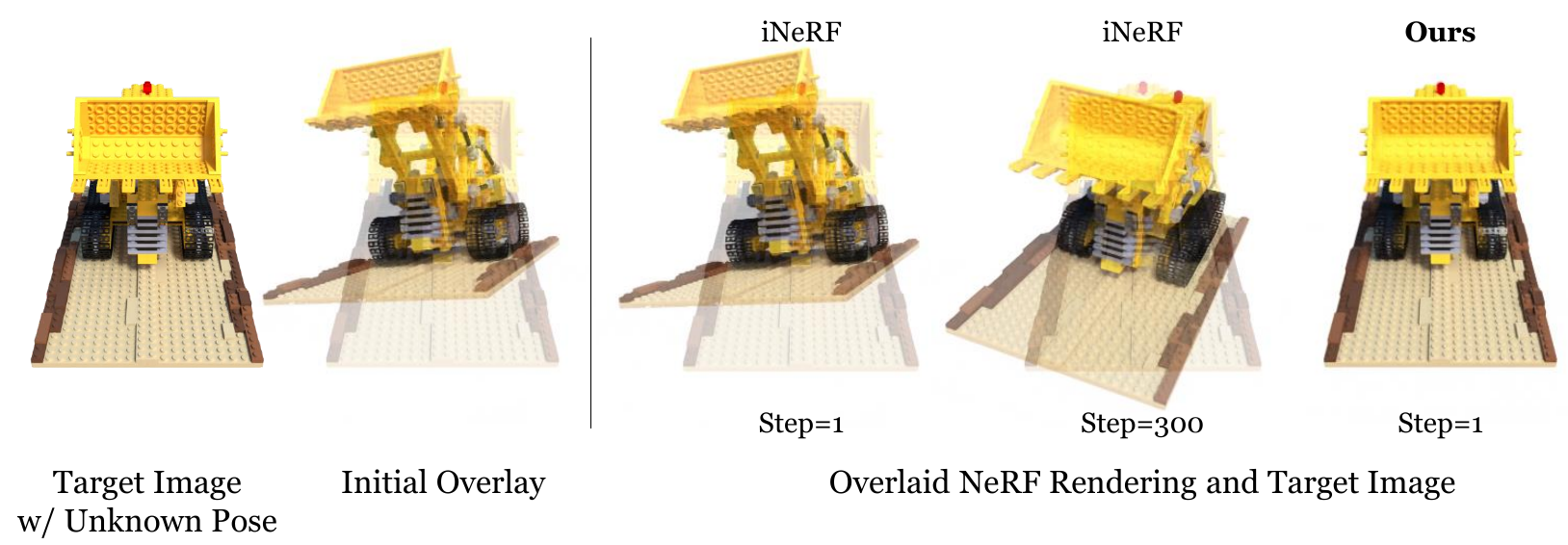}
		\captionof{figure}{Given an object image with unknown pose, we propose a NeRF-based pose estimation method, which reduces the hundreds of optimization steps in former NeRF-based method to only \textit{one} step, while avoiding being stuck in local minima, and obtaining more accurate poses. As a result, with only 5 minutes training of a fast NeRF~\cite{instant-ngp}, our method achieves CAD model-free real-time pose estimation on \textbf{\textit{novel}} objects at 6FPS.}
	\end{center}
}]
\renewcommand*{\thefootnote}{\fnsymbol{footnote}}
{
	\footnotetext[1]{The corresponding author is Prof. Yang Cong. The work is supported in part by National Key R\&D Program of China under Grant 2023YFB4704800, and NSFC under Grant 62225310, 62127807.}
}
\renewcommand*{\thefootnote}{\arabic{footnote}}

\begin{abstract}
Given the image collection of an object, we aim at building a real-time image-based pose estimation method, which requires neither its CAD model nor hours of object-specific training. Recent NeRF-based methods provide a promising solution by directly optimizing the pose from pixel loss between rendered and target images. However, during inference, they require long converging time, and suffer from local minima, making them impractical for real-time robot applications. We aim at solving this problem by marrying image matching with NeRF. With 2D matches and depth rendered by NeRF, we directly solve the pose in one step by building 2D-3D correspondences between target and initial view, thus allowing for real-time prediction. Moreover, to improve the accuracy of 2D-3D correspondences, we propose a 3D consistent point mining strategy, which effectively discards unfaithful points reconstruted by NeRF. Moreover, current NeRF-based methods naively optimizing pixel loss fail at occluded images. Thus, we further propose a 2D matches based sampling strategy to preclude the occluded area. Experimental results on representative datasets prove that our method outperforms state-of-the-art methods, and improves inference efficiency by 90$\times$, achieving real-time prediction at 6 FPS.
\end{abstract}


\section{Introduction}

Object pose estimation has wide applications in robot manipulation, augmented reality (AR) and mobile robotics~\cite{tcyb/CongCMLHY23}. Traditional methods typically require the CAD model of the object in advance, and searching for handcrafted features~\cite{iccv/Lowe99,hinterstoisser2011gradient} between the preregistered images or templatesand the target image. However, obtaining such high-quality CAD model can be difficult and labor-intensitive, or requires specialized high-end scanners. 
Recent methods have been applying deep neural network to regress the poses~\cite{xiang2018posecnn, wang2019densefusion,peng2019pvnet,di2022gpv,chen2020category}. However, they can only estimate poses of known instances~\cite{xiang2018posecnn, wang2019densefusion,peng2019pvnet} or similar ones from the same category~\cite{di2022gpv,chen2020category,lee2021category,chen2023stereopose}, and have to retrain on novel objects for hours. Moreover, they require large amount of training data, which is tedious to collect and annotate. Thus, it is difficult to apply such methods in real world due to unaffordable training time and human labor.

To further avoid tedious retraining for each novel object, recent methods~\cite{sun2022onepose, he2022oneposeplusplus} learn from the traditional pipeline of SfM (Struction-from-Motion) to estimate object poses via feature matching. Given a small set of multi-view images, they first reconstruct sparse point cloud of the object via SfM, and then form 2D-3D correspondences to estimate the pose by solving the PnP~\cite{EPnP} problem. Unfortunately, such methods rely on forming stably repeatable correspondences across all input frames, which usually cannot be guaranteed, thus leading to large pose error. 
On the other hand, recent advances in NeRF (Neural Radiance Fields~\cite{nerf,inerf,lin2023parallel, instant-ngp}) provide a mechanism for capturing complex 3D geometry in a few minutes. Following former render-and-compare methods for pose estimation~\cite{park2020latentfusion, palazzi2018end, chen2019learning}, iNeRF~\cite{inerf} first trains a NeRF from image collection, and then during testing, it optimizes the pose by minimizing dense pixel error between the rendered and target image. Such dense supervision allows iNeRF to achieve more accurate alignment, but it also requires hundreds of iterations taking minutes. Moreover, its convergence relies on good initialization, and typically fails at large pose differences or occlusion.

In this work, we try to combine the best of both worlds by marrying image matching with NeRF to achieve \textit{real-time} image-based pose estimation, without hundred steps of optimization. With 2D pixel matches and corresponding depth rendered by NeRF, we can build 2D-3D correspondences, and directly solve the pose with PnP~\cite{EPnP}. This significantly reduces the iteration number and allows for real-time inference for NeRF based method. Moreover, comparing to former keypoint-based method~\cite{sun2022onepose,he2022oneposeplusplus}, this eases the difficulty of building 2D-3D correspondences in traditional SFM-based methods, which needs to find 2D matches between multiple input frames and the target image. With NeRF, our method only matches between two images once, and can convert arbitrary 2D matches to 2D-3D correspondences by backprojecting NeRF rendered depth into 3D space. 

Moreover, owing to the implicit nature of NeRF, the rendered depth can be noisy and unfaithful~\cite{yariv2021volume,yariv2020multiview,neus}. To improve the quality of 2D-3D correspondences, we further propose a 3D consistent point mining strategy to discard unfaithful and noisy 3D points reconstructed by NeRF so that the PnP can obtain more accurate poses. Specifically, we render the 3D points from nearby viewpoints and regard the variation of them as the 3D consistency. 

Our method also allows for further pose refinement from pixel error, like former render-and-compare methods~\cite{park2020latentfusion, palazzi2018end,inerf}. However, this process is sensitive to occlusion, which bakpropagtes false gradient to the pose. We notice that the matching points indicate unoccluded area, and propose a matching point based sampling strategy for loss computation. 
In experiments, we show that our proposed method improves the efficiency over former NeRF based methods by \textbf{90 times}, and can inference in real-time at 6FPS, while achieving higher pose accuracy and stronger robustness to occlusion.

Our contributions are three folds:
1) An efficient NeRF based pose estimation method is proposed by introducing image matching, which allows real-time image-based inference, and is free of CAD model or hours of pretraining.
2) We propose a 3D consistent point mining strategy to detect and discard unfaithful points reconstructed by NeRF to enable more accurate pose estimation.
3) In contrast to former render-and-compare based methods, our method can overcome the occlusion problem with a matching point based sampling strategy.

\section{Related Works}
\subsection{Deep Learning Based Pose Estimation}

Recently, deep neural networks have led a series of breakthroughs for pose estimation. Some methods\cite{xiang2018posecnn, wang2019densefusion, kehl2017ssd} focus on regressing the object pose directly. Some methods\cite{rad2017bb8, tekin2018real, song2020hybridpose} train neural networks to build the 2D-3D correspondence first and then apply the PnP algorithm to compute the 6-DoF poses. OSOP~\cite{shugurov2022osop} proposes a one-shot method by first using a textured 3D template to match target image, and then solve the pose from dense 2D-3D correspondences constructing by image matching. Recently, some methods \cite{di2022gpv,chen2020category,lee2021category,chen2023stereopose} leverage category-level representation to estimate both the pose and the scale of novel instances within the same category. Although great success has been made, they have to either obtain high-quality CAD models or spend expensive costs to collect and annotate large amount of data, severely limiting their application in the real world.

\subsection{Render-and-compare Based Pose Estimation}

\noindent
\textbf{Traditional} render and compare methods~\cite{palazzi2018end,chen2019learning,megapose,ponimatkin2022focal} first render the 3D CAD model and compare with input 2D images, and then minimize the error to optimize the pose. So they typically require high-quality 3D models in advance, which cannot be applied in our CAD-free setting. Though 3D models can be obtained from multi-view images via differentiable renders~\cite{liu2019soft,petersen2022gendr,park2020latentfusion}, the reconstruction and rendering quality is limited, and may fail at complex real-world scenes.

\noindent
\textbf{NeRF-based.}
Neural Radiance Fields~\cite{nerf} provide a remedy for render-and-compare based strategy with its remarkable improvement in rendering quality. INeRF~\cite{inerf} first proposes to estimate the pose by inverting NeRF, \textit{i.e.}, optimizing the pose from image difference. Though achieving accurate results, it requires hundreds of iterations, and struggles at converging to correct poses. To solve these problems, Loc-NeRF~\cite{maggio2023loc} and~\cite{lin2023parallel} propose to use Monte Carlo sampling to improve the efficiency and robustness to local minima. However, these methods still require optimization, thus cannot achieve real-time estimation. On the other hand, NeRF has been used in SLAM methods~\cite{sucar2021imap, zhu2022nice,tosi2024nerfs}, where the camera poses are required to be estimated. iMap~\cite{sucar2021imap} and NICE-SLAM~\cite{zhu2022nice} use depth captured by RGB-D camera to supervise the localization process. In contrast, our method aims at estimating pose from RGB images.

\subsection{Keypoint-Matching-Based Pose Estimation}
Traditional methods~\cite{tang2012textured,martinez2010moped,collet2009object} use hand-crafted features like SIFT\cite{lowe2004distinctive}, FAST\cite{rosten2006machine} and ORB\cite{rublee2011orb} to match interest points between training images and a pre-built 3D model to solve the pose. Nowadays, some methods introduce deep learning to improve accuracy of matches. \cite{yi2018learning, ranftl2018deep} train a classifier to distinguish inliers and outliers. SuperGlue\cite{sarlin2020superglue} designs self- and cross-attention layers to enhance the exploration of features relations. Recently, OnePose\cite{sun2022onepose} assigns features to SfM reconstructed 3D points by aggregating image features from multiple training views, and directly matches with target images. Onepose++~\cite{he2022oneposeplusplus} later improves it with a keypoint-free reconstruction framework. However, they still rely on large training data to train feature aggregation networks limiting their application.

\section{Background}
\textbf{NeRF.} Given multi-view images with annotated camera parameters, NeRF~\cite{nerf} represents scenes via a 5D function:
\begin{equation}\label{eq:nerf}
\mathbf{c}, \sigma=\Phi(\mathbf{x}, \mathbf{d}),
\end{equation}
which maps the query point location $\mathbf{x}\in \mathbb{R}^3$ to its density $\sigma\in \mathbb{R}^1$, and view-dependent color $\mathbf{c}\in \mathbb{R}^3$ at direction $\mathbf{d}\in \mathbb{R}^3$. It reconstructs the scene implicitly, and is able to render freeview images. To render an image from view $P$,
the color $\hat{C}(\mathbf{p},P)$ of a pixel $\mathbf{p}\in \mathbb{R}^2$ is obtained by accumulating the color along rays $\mathbf{r}$ that passes the pixel, following the volume rendering technique~\cite{kajiya1984ray}:
\begin{equation}
\hat{C}(\mathbf{p},P)=\sum_{i=i}^N \omega_i \mathbf{c}_i,
\end{equation}
where $\omega_i=\sum_{i=i}^N T_i(1-\exp(\sigma_i\delta_i))$ is the weight of each ray point, $T_i=\exp(-\sum_{j=1}^{i-1}{\sigma_j\delta_j})$, and $\delta_i$ is the sample step along the ray. Similarly, we can also render an approximate depth at pixel $\mathbf{p}$ by
\begin{equation}\label{eq:depth}
\hat{z}(\mathbf{p},P) = \sum_{i=i}^N\omega_i t_i,
\end{equation}
where $t_i\in \mathbb{R}^1$ is the depth at each ray point.
\begin{figure*}
	\centering
	\includegraphics[width=0.92\textwidth]{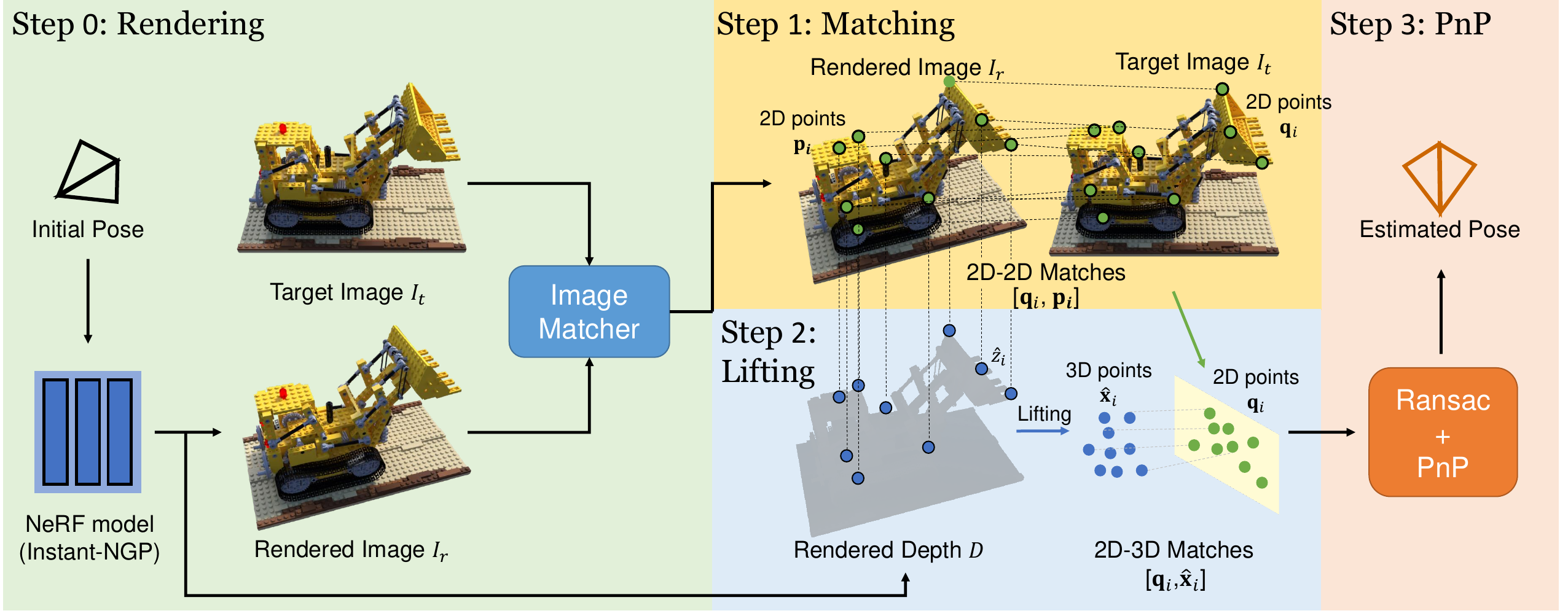}
	\caption{Framework of the one-step pose estimation via feature matching strategy. Given the initial pose, we use NeRF~\cite{instant-ngp} to render an RGB image $I_r$, and a depth image $D$. Then, an off-the-shelf image matcher~\cite{loftr} is applied to generate 2D-2D matches between the rendered and target image. Given location of matched 2D points and its depth rendered by NeRF, the 3D coordinates can be obtained, thus forming 2D-3D matches, from which the pose is finally solved via PnP+RANSAC.}\label{fig:main}
 	\vspace{-15pt}
\end{figure*}

{\bf InstantNGP.}
The original NeRF suffers from tediously lengthy training, and is infeasible to run in real time. InstantNGP~\cite{instant-ngp} improves the efficiency by decomposing the scene into a multi-resolution hash table and tiny MLP. It significantly reduces the training time to 5min, and allows for real-time rendering. For application in pose estimation, this makes fast online training of novel objects, and real-time inference possible. So we use it as default NeRF model.

{\bf NeRF-based Pose Estimation.} INeRF first proposes to estimate the pose of a novel object with NeRF. It first trains a NeRF model $\Phi$ with multi-view images of the object. Then, during inference, given a new target image $I_t$, iNeRF~\cite{inerf} recovers the camera pose $T \in SE(3)$ by optimizing:
\begin{equation}~\label{eq:inerf-func}
\hat{T}=\underset{T \in \operatorname{SE}(3)}{\operatorname{argmin}}\, \|\Phi(T)-I_t\|_2,
\end{equation}
where $\Phi(T)$ denotes NeRF rendered image from view $T$, and the function denotes an L2 loss between $\Phi(T)$ and the target image $I_t$. The NeRF weights are fixed in optimization. 

\section{Method}
Our method aims at improving the convergence speed of NeRF-based pose estimation method. The key insight is to marry feature matching with NeRF to directly solve the pose from 2D-3D correspondences via PnP, which we introduce in~\ref{sec:pnp}. Moreover, owing to the implicit nature of NeRF, 3D coordinates lifted from 2D pixels can be noisy and unfaithful. Thus, in Sec.~\ref{sec:consisloss}, we improve the 3D consistency by introducing a \textit{3D consistent point mining} strategy before solving the pose. So far, without any refinement, our result is already more accurate than iNeRF~\cite{inerf} in most cases, which needs hundreds steps of refinement. Our method also allows further optimization to refine the initial pose. However, we notice that current pixel error (Eq.~\ref{eq:inerf-func}) cannot handle occluded images. For this, we propose a keypoint-guided occlusion robust refinement to tackle the occlusion problem, which is introduced in Sec.~\ref{sec:kps}.

\subsection{One-step Pose Estimation via Feature Matching}\label{sec:pnp}
Optimizing the pose from the photometric loss between rendered and target image following the formulation of iNeRF~\cite{inerf} (Eq.~\ref{eq:inerf-func}) can be extremely challenging, due to highly non-convex objective function. As a result, current methods are prone to being stuck in local minima. Here, we propose to estimate the pose by marrying image matching with NeRF. As shown in Fig.~\ref{fig:main}, the method has three main steps: 
\subsubsection{Matching}
To estimate the pose of the target image $I_t$, we first render an image $I_r$ from the initial guess of camera pose $P$ with the trained NeRF model. Then, a pretrained off-the-shelf image matching model~\cite{loftr,sarlin2020superglue,sarlin2021back} is applied to form 2D-2D matches $[\mathbf{q}_i, \mathbf{p}_i]$ between the target image $I_t$ and the rendered image $I_r$, with $\mathbf{q}_i\in I_t$ and $\mathbf{p}_i\in I_r$. We apply the recent proposed transformer-based image matching method LoFTR~\cite{loftr} in all our experiments. 
\subsubsection{Lifting}~\label{lift}
We then convert 2D-2D matches $[\mathbf{q}_i, \mathbf{p}_i]$ between target image $I_t$ and rendered image $I_r$ to 2D-3D correspondences $[\mathbf{q}_i, \mathbf{x}_i]$. We achieve this by lifting the matched 2D pixels $\mathbf{p}\in  \mathbb{R}^2$ in NeRF rendered image $I_r$ to 3D space. Specifically, we first obtain the depth $\hat{z}_i$ from the depth map $D$ rendered by the trained NeRF model following Eq.~\ref{eq:depth}. Then, the 3D coordinate of the corresponding point $\hat{\mathbf{x}}$ is obtained via backprojection, and transformed to world space via current camera pose $P$:
\begin{equation}
	\hat{\mathbf{x}}_i=P\hat{z}_iK^{-1}\mathbf{p}_i
\end{equation}

\subsubsection{PnP} After obtaining the 2D-3D correspondences, the pose is computed via PnP~\cite{EPnP} with RANSAC~\cite{ransac}. The above procedure already allows us to obtain good pose with only one rendering step, which is much faster than former NeRF based baselines~\cite{inerf, lin2023parallel}. However, there may exists error due to inaccurate feature matches. In the following, we introduce a strategy to further improve the performance. 

\subsection{3D Consistent Point Mining}\label{sec:consisloss}
In the above framework, one of the key factors that affect the pose accuracy is the precision of the 2D-3D matches, which are computed by a trained NeRF~\cite{nerf} model as stated in~\ref{lift}. However, owing to the implicit nature of NeRF, the learned scene geometry can be unfaithful and noisy~\cite{yariv2021volume,yariv2020multiview,neus}. Moreover, the estimated 3D coordinates can be inconsistent when rendering from different views, resulting in large pose error. These problems become severer when the training images are limited, or the camera poses are noisy. 

To counter the above problem, we propose to preclude the inconsistent 3D points by introducing a 3D consistent point mining strategy. Specifically, for each 3D keypoint $\mathbf{x}$ that is lifted from a matched 2D pixel $\mathbf{p}$, its consistency $m$ is evaluated by re-estimating the 3D coordinates from nearby views, and computing how well these points are aligned with each other. 

Specifically, given the current view $P$ and estimated 3D keypoint $\mathbf{x}$, we first sample $k$ nearby views $\mathcal{P} =\{P_i\}_{i=1}^k$. Then, we shoot rays $R=\{\mathbf{r}_i\}_{i=1}^k$ that pass the 3D keypoint $\mathbf{x}$ from each view in $\mathcal{P}$, and estimate the 3D coordinates $X=\{\mathbf{x}_i\}_{i=1}^k$ on these rays:
\begin{equation}
X=\hat{Z}\cdot \operatorname{norm}({\mathcal{P}}P^{-1}\mathbf{x}),
\end{equation}
where $\hat{Z}=\{\hat{z}_i\}_{i=1}^k$ is the depth value of rays $R$ estimated by NeRF, and $\operatorname{norm}(\cdot)$ denotes vector normalization.
We measure the point consistency with the location variance:
\begin{equation}
m=\frac{1}{k}||X-\mathbf{x}||_2^2,
\end{equation}
where larger $m$ indicates lower consistency. Finally, we introduce a threshold $\gamma$ to discard the points whose consistency $m>\gamma$, where $\gamma$ is determined empirically.

\subsection{Keypoint-guided Occlusion Robust Refinement}~\label{sec:kps}
Current NeRF-based method cannot estimate the pose of occluded images. The reason is that the photometric loss computed from occluded area will backpropagate false gradients to the pose, which will aggravate the issue of being stuck in local minima. 

Our image-matching based strategy provides a solution to this problem. Assuming the image matcher to be accurate enough, the matched keypoints naturally provide cues for unoccluded area, thus preventing the false gradients. We propose to compute the photometric loss with a new  matched keypoint-guided sampling strategy. Specifically, after predicting matches, we apply $5\times 5$ morphological dilation around the matched keypoint for $n$ times to obtain the sample region.

\section{Experiments}
We evaluate the pose estimation performance of our proposed method on NeRF synthetic dataset~\cite{nerf} and complex real-world scene from LLFF dataset~\cite{llff}.

\subsection{Comparison Methods}
We evaluate our method by comparing against state-of-the-art NeRF based pose estimation methods, and image matching based method:

\textit{\textbf{iNeRF}~\cite{inerf}} is the first method to estimate object poses by inverting neural radiance fields. It computes photometric loss between the rendered and target images, and backpropagate through NeRF's framework to optimize the pose. 

\textit{\textbf{pi-NeRF}~\cite{lin2023parallel}} improves the efficiency of iNeRF by using instant-NGP~\cite{instant-ngp}. It overcomes the local minimum by parallelly optimizing and pruning Monte Carlo sampled poses.

\textit{\textbf{LoFTR~\cite{loftr}} }, where we directly solve the pose from 2D matches estimated by LoFTR via epipolar geometry. The translation evaluation is omitted due to scale ambiguity.

\textit{\textbf{Ours (1-step)}} To demonstrate the significance of the proposed feature matching strategy, we build Ours (1-step) baseline. It takes the PnP solved pose as final results, and does not apply further pose refinement.

\subsection{Results on Synthetic Dataset}\label{sec:syn_data}
\subsubsection{Setting} We choose Instant-ngp~\cite{instant-ngp} as the NeRF model, and train it on all the training images. For evaluation, we follow iNeRF~\cite{inerf} to choose 5 test images from test set to estimate the pose. For each image, 5 initial poses are sampled by rotating around a random axis by a random angle within $[10^{\circ}, 40^{\circ}]$, and translating along a random vector by length within 0.2. To explore the performance limits, such initial pose perturbation is severer than former work~\cite{inerf,lin2023parallel}, so the results of the comparison methods may be worse than the results reported in the original paper. All comparison methods except for iNeRF\footnote{https://github.com/salykovaa/inerf} are evaluated with the official implementation.

\begin{table}[t] 
	\vspace*{0.06in}
	\caption{6-DoF pose estimation Results on the NeRF Synthetic and LLFF datasets, where RE / TE denote rotation / translation error, respectively. mRE / mTE denote mean rotation / translation error over all subjects.
		\label{tab:synthetic_and_llff}}  
	\centering 
	\begin{tabular}{lcccc}
		\toprule
		Method                &  RE$<$$5^{\circ}$($\uparrow$) & TE$<$$0.05$($\uparrow$) & mRE ($\downarrow$) & mTE ($\downarrow$) \\
		\cmidrule(r){1-5}
		\multicolumn{5}{l}{\textbf{NeRF Synthetic Dataset}} \\
		\cmidrule(r){1-5}
		
		iNeRF~\cite{inerf}           &0.585   & 0.56& 10.33  &0.559  \\
		pi-NeRF~\cite{lin2023parallel}   & 0.24  & 0.04  & 15.83  & 1.073   \\
		LoFTR~\cite{loftr}    & 0.785  & -  & 6.15  &  -  \\
		Ours (1-step)   & 0.945  & 0.75  & 1.57  &  0.096  \\
		Ours & {\bf 0.95}     &  {\bf 0.88}      &  {\bf 1.25}      &  {\bf 0.077}        \\
		\cmidrule(r){1-5}
		\multicolumn{5}{l}{\textbf{LLFF Dataset}} \\
		\cmidrule(r){1-5}
		
		iNeRF~\cite{inerf}           & 0.50  & 0.55  & 16.46  & 0.0618  \\
		pi-NeRF~\cite{lin2023parallel} &  0.00 &  0.00 & 133.37  & 3.999 \\
		LoFTR~\cite{loftr}    & 0.994  & -  & 0.667  &  -  \\
		Ours (1-step)   & \textbf{1.00}  & {\bf 1.00}  & 0.325  &  \textbf{0.0027} \\
		Ours &  {\bf 1.00}  &  {\bf 1.00}  & \textbf{0.135}     & \textbf{0.0008}  \\
		\bottomrule
	\end{tabular}
	\vspace*{-10 pt}
\end{table} 

\subsubsection{Results}
As shown in Tab.~\ref{tab:synthetic_and_llff}, we report the pose correctness, \textit{i.e.}, the rate of poses with rotation error $<5^{\circ}$, and translation error $<5$ units, and mean rotation (mRE) and translation error (mTE).

\begin{figure}
	\includegraphics[width=0.48\textwidth]{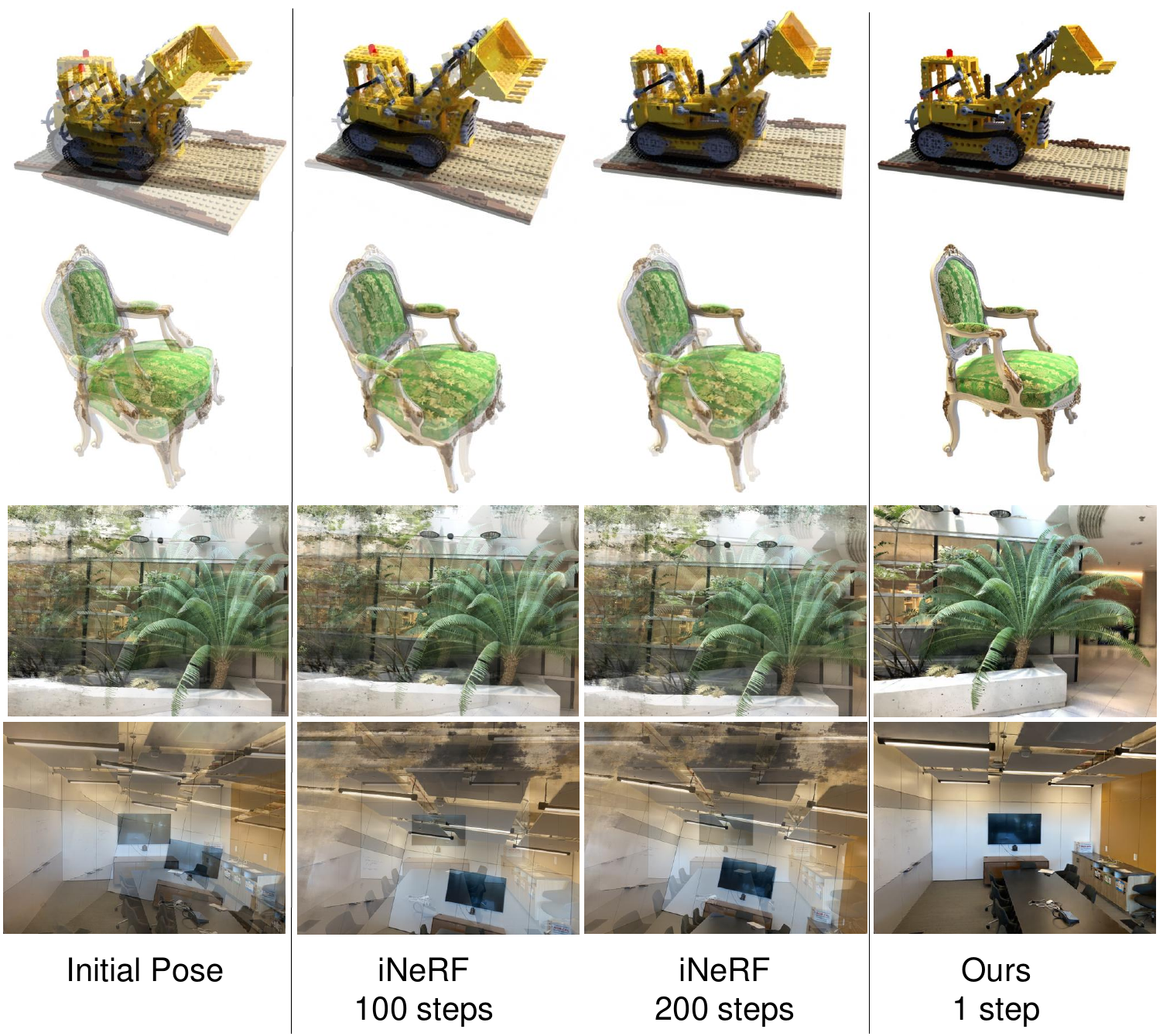}
	\captionof{figure}{Qualitative results of pose estimation on NeRF synthetic~\cite{nerf} and real-world LLFF dataset~\cite{llff}. We visualize the results by overlying the target image and NeRF rendering image from the estimated pose.} \label{fig:qual}
	\vspace{-10pt}
\end{figure}

On NeRF Synthetic dataset, \textit{Ours (1-step)} already outperforms NeRF-based methods by 36$\%$ and 19$\%$ in terms of the rotation and translation accuracy. Moreover, pi-NeRF~\cite{lin2023parallel} achieves worse performance that iNeRF~\cite{inerf}. We assume the reason is that pi-NeRF fails to guess good initial pose under such severe pose perturbation, and abandoning the interest region based pixel loss used in iNeRF makes the convergence even harder. Our method is also superior than direct solving pose from LoFTR~\cite{loftr} 2D matches via epipolar geometry, which indicating that our idea of combining 2D matches with 3D information provided by NeRF can complement each other, and further boost the performance. With post refinement of 40 steps, our full method can further boost the correctness of rotation and translation from 94.5$\%$ / 75$\%$ to 95$\%$ / 88$\%$.
Such few steps of optimization is much less than iNeRF (300 steps), and pi-NeRF (2500 steps), thanks to the accurate initial pose obtained by 1-step pose solving strategy. It can effectively alleviates the local minima suffered by pure optimization based method. The qualitative results shown in Fig.~\ref{fig:qual} shows that our method achieves nearly perfect alignment under large initial pose differences.

\subsection{Results on Real World Scene}
\subsubsection{Setting} We evaluate on 4 complex scenes captured by LLFF~\cite{llff} including \textit{Fern}, \textit{Fortress}, \textit{Horns}, and \textit{Room} following iNeRF. The model and protocol for pose initialization is the same as in iNeRF~\cite{inerf}. Here, the scenes are captured from forward view. So the setting is closer to the visual localization task in SLAM.

\subsubsection{Results}
On real-world scene, similar to NeRF Synthetic dataset, our method achieves the best results. This dataset is more challenging, because the scenes are captured with forward-facing images, which will result in larger image differences under the same rotation angle. As a result, it leads to performance degradation for the comparison methods. On the contrary, our method even achieves better results (100$\%$). This verifies the robustness of our method to large pose variations. Compared to synthetic dataset, the improved performance may be because the matcher~\cite{loftr} performs better on real-world data.
\begin{figure*}
	\centering
	\includegraphics[width=0.94\textwidth]{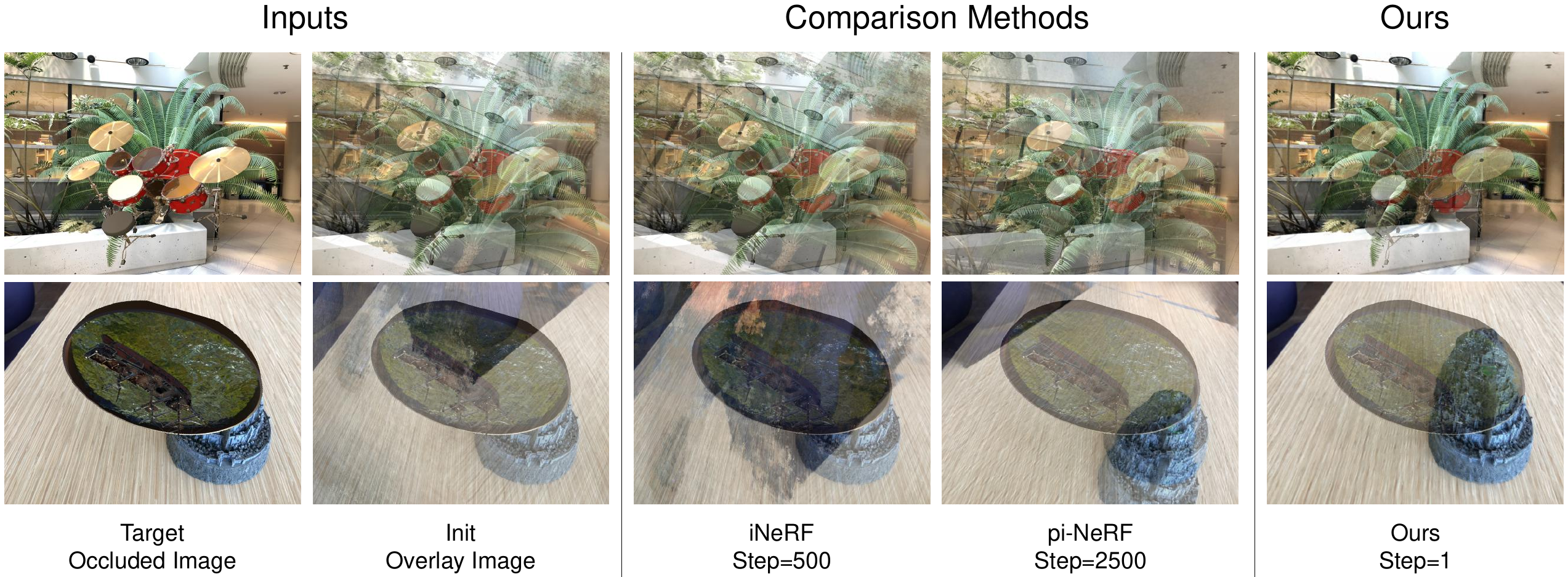}
 	\vspace{-5pt}
	\captionof{figure}{Qualitative results of pose estimation on synthesized occluded data. The comparison methods fail to align the occluded images after hundreds of iterations, while our method aligns well in one step.} \label{fig:occ}
 	\vspace{-10pt}
\end{figure*}

\subsection{Results on Occluded Dataset}

\subsubsection{Setting} We further explore the performance of our proposed method on occluded dataset. The occluded data is synthesized by composing the NeRF synthetic and LLFF dataset. The LLFF real-world images are used as background, and objects from synthetic dataset are randomly transformed and added as foreground. See the leftmost column of Fig.~\ref{fig:occ} for an example.

\subsubsection{Results} We show qualitative pose estimation results on occluded dataset in Fig.~\ref{fig:occ}. For \textbf{iNeRF}~\cite{inerf}, taking fortress (second row) as an example, with the main object being occluded, the photometric loss computed from repeated pattern of the table cannot guide the pose optimization.
For \textbf{pi-NeRF}~\cite{lin2023parallel}, Monte Carlo sampling strategy also fail on occluded images, because the photometric error from occluded area can no longer be used as a criterion to sample correct pose. 
On the contrary, \textbf{our method} bypasses the occluded area by leveraging the power of matching method~\cite{loftr}, and still solves good pose in one step. The effect of occlusion robust matching area guided sampling strategy is analyzed in Tab.~\ref{tab:ablation} of ablation study .

\subsection{Efficiency}
Efficiency is one of our key advantages. We evaluate the efficiency on NeRF Synthetic dataset, and run all experiments on one RTX3090 GPU. Without post refinement, our method runs at 6FPS, and is \textbf{90$\times$} faster than former best method~\cite{lin2023parallel}, including $\sim$50ms for rendering, $\sim$60ms for matching, $\sim$15ms for PnP+RANSAC. For post refinement, our method requires much less iterations (40) comparing to iNeRF~\cite{inerf} (300) and pi-NeRF~\cite{lin2023parallel} (2500), which also boost the efficiency of pose refinement. We attribute the reduction to the good initialization of our keypoint based strategy. Optimizing Instant-NGP~\cite{instant-ngp} for 40 steps takes about 400ms.

\subsection{Ablation Studies}
\begin{table}[t]
	\centering
	\caption{Results of ablation studies on the proposed 3D Consistent Points Mining Strategy (3D Consis.), and Keypoint-guided Occlusion Robust Refinement (KOR) strategy, where rot. and trans. denote rotation and translation.}
	\label{tab:ablation}
	\resizebox{0.45\textwidth}{!}{
		\begin{tabular}{@{~}l@{~}l|c|c}
			\toprule
			& Balines & Mean rot. error ($^\circ$) $\downarrow$ & Mean trans. error (cm) $\downarrow$ \\ \toprule
			&& \multicolumn{2}{c}{NeRF Synthetic Dataset} \\
			I & w/o 3D Consis. (1-step)  & 2.08  &  0.133    \\
			II & w/ 3D Consis. (1-step) & \textbf{1.57} & \textbf{0.096}  \\
			\midrule
			&	& \multicolumn{2}{c}{Occluded LLFF Dataset} \\
			III & w/o KOR Refine. & 0.562 & 0.33   \\
			IV & w/ KOR Refine & \textbf{0.518} & \textbf{0.29}   \\
			\bottomrule
		\end{tabular}%
	}
\end{table}

We explore the effectiveness of each proposed components, including the 3D consistent point mining strategy and the keypoint-guided sampling strategy for occlusion robust pose refinement. 

\begin{figure}
	\centering
\vspace{-5pt}
	\includegraphics[width=0.48\textwidth]{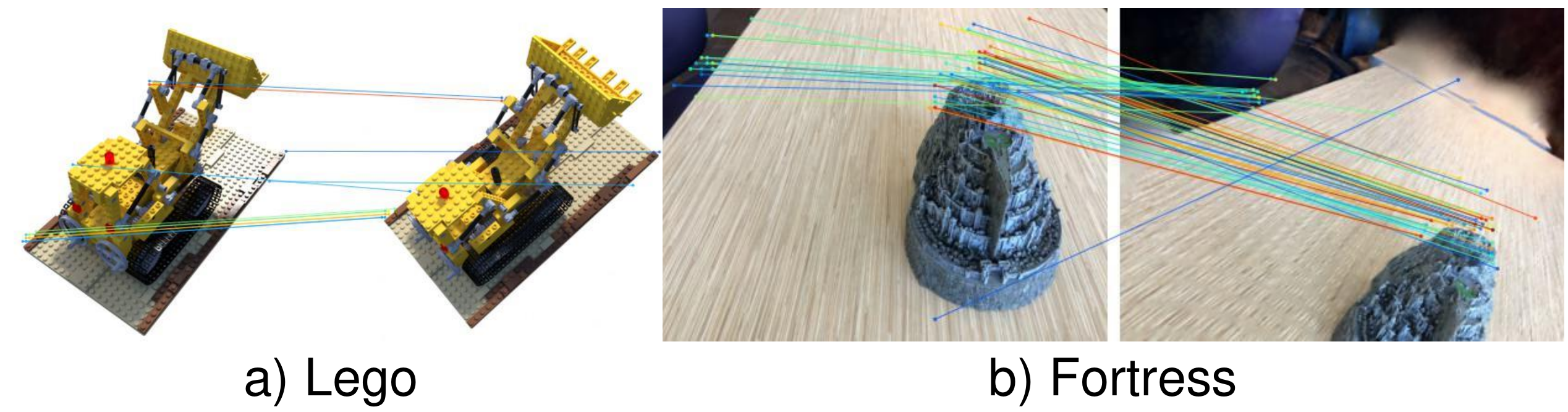}
	\caption{Visualization of the points discarded by 3D consistent point mining strategy. } \label{fig:ablation-consis}
 	\vspace{-10pt}
\end{figure}

\textit{\textbf{3D Consistent Point Mining.}} To explore how 3D consistent point mining works, we first visualize the inconsistent points that are discarded in Fig.~\ref{fig:ablation-consis}. In the Lego images, the inconsistent points typically locate near the silhouette, where NeRF cannot reconstruct well, and slight mismatch of pixels may cause large change of depth. Similarly, in the fortress image, the discarded points also appear near the fortress edge. Other discarded points are at the corner of the image, whose geometry is also not well-defined as they are rarely seen. In conclusion, the proposed 3D consistent point mining strategy can automatically detect and discard unstable 3D points learned by NeRF, thus improving the pose accuracy. 

We also evaluate 3D Consistent Point Mining strategy quantitatively on NeRF synthetic dataset in Tab.~\ref{tab:ablation}. Here, we report the 1-step results without post refinement. The results validate its effectiveness.

\textit{\textbf{Keypoint-guided Occlusion Robust Refinement.}} 
\begin{figure}
	\centering
	\includegraphics[width=0.48\textwidth]{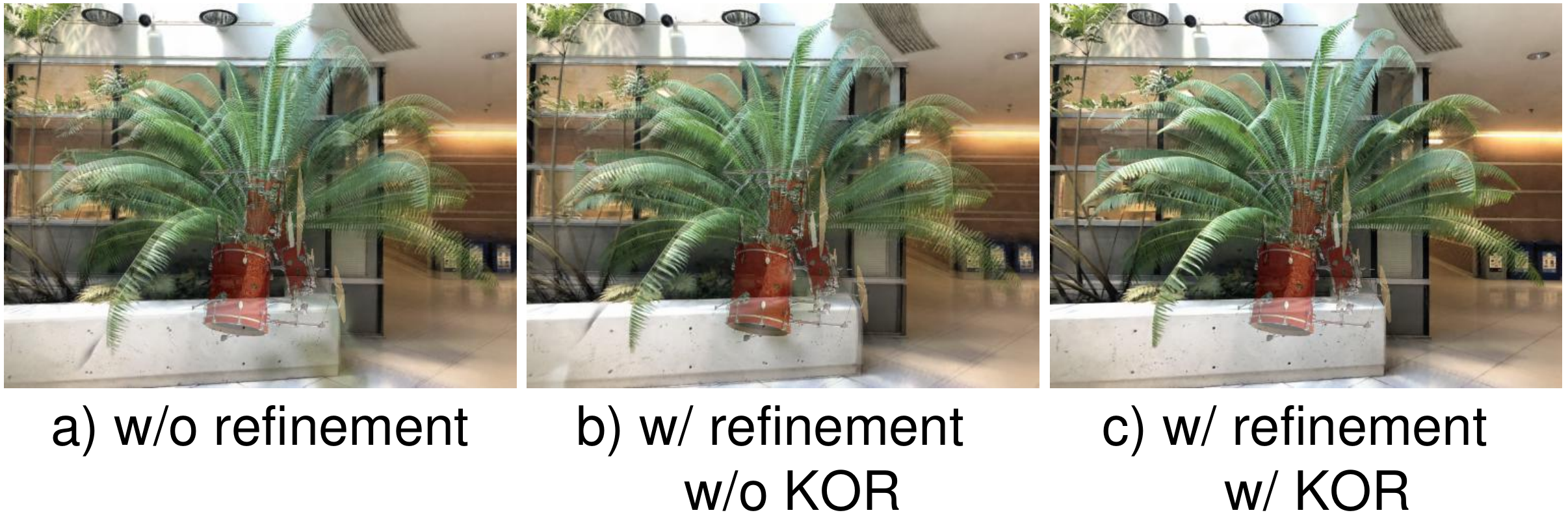}
 	\vspace{-15pt}
	\caption{Visualization of the effect of the proposed keypoint-guided Occlusion Robust Refinement strategy (KOR). } \label{fig:ablation-KOR}
   	\vspace{-10pt}
\end{figure}
As shown in Fig.~\ref{fig:ablation-KOR}, the proposed \textit{keypoint-guided occlusion robust} (KOR) strategy achieves accurate alignment under severe occlusion, while the interest area sampling strategy~\cite{inerf} (w/refinement, w/o KOR) suffers from local minima. Quantitatively, as shown in Tab.~\ref{tab:ablation}, KOR reduces rotation and translation error by $4.4\%$.

\section{Conclusion}
We have proposed a fast NeRF-based framework for imaged-based, CAD-free novel object pose estimation. By introducing keypoint matching, our method can directly solve the pose with one step, and is free of long optimization time and local minima. Moreover, we propose a 3D consistent point mining strategy to improve the quality of 2D-3D correspondences, and a matching keypoint based sampling strategy to improve the robustness to occluded images. Experiments demonstrate our superior performance and robustness to occlusion. For future work, we hope that this method can be extended to robot manipulation or recent neural field based SLAM tasks~\cite{yugay2023gaussian, matsuki2023gaussian,yan2023gs,keetha2023splatam,tosi2024nerfs} to push the efficiency limit of localization. 

\bibliographystyle{IEEEtran}
\bibliography{IEEEfull}

\begin{thebibliography}{10}
\providecommand{\url}[1]{#1}
\csname url@samestyle\endcsname
\providecommand{\newblock}{\relax}
\providecommand{\bibinfo}[2]{#2}
\providecommand{\BIBentrySTDinterwordspacing}{\spaceskip=0pt\relax}
\providecommand{\BIBentryALTinterwordstretchfactor}{4}
\providecommand{\BIBentryALTinterwordspacing}{\spaceskip=\fontdimen2\font plus
\BIBentryALTinterwordstretchfactor\fontdimen3\font minus
  \fontdimen4\font\relax}
\providecommand{\BIBforeignlanguage}[2]{{%
\expandafter\ifx\csname l@#1\endcsname\relax
\typeout{** WARNING: IEEEtran.bst: No hyphenation pattern has been}%
\typeout{** loaded for the language `#1'. Using the pattern for}%
\typeout{** the default language instead.}%
\else
\language=\csname l@#1\endcsname
\fi
#2}}
\providecommand{\BIBdecl}{\relax}
\BIBdecl

\bibitem{instant-ngp}
T.~M\"uller, A.~Evans, C.~Schied, and A.~Keller, ``Instant neural graphics
  primitives with a multiresolution hash encoding,'' \emph{ACM Trans. Graph.},
  vol.~41, no.~4, pp. 102:1--102:15, Jul. 2022.

\bibitem{tcyb/CongCMLHY23}
Y.~Cong, R.~Chen, B.~Ma, H.~Liu, D.~Hou, and C.~Yang, ``A comprehensive study
  of 3-d vision-based robot manipulation,'' \emph{{IEEE} Trans. Cybern.},
  vol.~53, no.~3, pp. 1682--1698, 2023.

\bibitem{iccv/Lowe99}
D.~G. Lowe, ``Object recognition from local scale-invariant features,'' in
  \emph{Proceedings of the International Conference on Computer Vision,
  Kerkyra, Corfu, Greece, September 20-25, 1999}.\hskip 1em plus 0.5em minus
  0.4em\relax {IEEE} Computer Society, 1999, pp. 1150--1157.

\bibitem{hinterstoisser2011gradient}
S.~Hinterstoisser, C.~Cagniart, S.~Ilic, P.~Sturm, N.~Navab, P.~Fua, and
  V.~Lepetit, ``Gradient response maps for real-time detection of textureless
  objects,'' \emph{IEEE transactions on pattern analysis and machine
  intelligence}, vol.~34, no.~5, pp. 876--888, 2011.

\bibitem{xiang2018posecnn}
Y.~Xiang, T.~Schmidt, V.~Narayanan, and D.~Fox, ``Posecnn: A convolutional
  neural network for 6d object pose estimation in cluttered scenes,''
  \emph{Robotics: Science and Systems XIV}, 2018.

\bibitem{wang2019densefusion}
C.~Wang, D.~Xu, Y.~Zhu, R.~Mart{\'\i}n-Mart{\'\i}n, C.~Lu, L.~Fei-Fei, and
  S.~Savarese, ``Densefusion: 6d object pose estimation by iterative dense
  fusion,'' in \emph{Proceedings of the IEEE/CVF conference on computer vision
  and pattern recognition}, 2019, pp. 3343--3352.

\bibitem{peng2019pvnet}
S.~Peng, Y.~Liu, Q.~Huang, X.~Zhou, and H.~Bao, ``Pvnet: Pixel-wise voting
  network for 6dof pose estimation,'' in \emph{Proceedings of the IEEE/CVF
  Conference on Computer Vision and Pattern Recognition}, 2019, pp. 4561--4570.

\bibitem{di2022gpv}
Y.~Di, R.~Zhang, Z.~Lou, F.~Manhardt, X.~Ji, N.~Navab, and F.~Tombari,
  ``Gpv-pose: Category-level object pose estimation via geometry-guided
  point-wise voting,'' in \emph{Proceedings of the IEEE/CVF Conference on
  Computer Vision and Pattern Recognition}, 2022, pp. 6781--6791.

\bibitem{chen2020category}
X.~Chen, Z.~Dong, J.~Song, A.~Geiger, and O.~Hilliges, ``Category level object
  pose estimation via neural analysis-by-synthesis,'' in \emph{Computer
  Vision--ECCV 2020: 16th European Conference, Glasgow, UK, August 23--28,
  2020, Proceedings, Part XXVI 16}.\hskip 1em plus 0.5em minus 0.4em\relax
  Springer, 2020, pp. 139--156.

\bibitem{lee2021category}
T.~Lee, B.-U. Lee, M.~Kim, and I.~S. Kweon, ``Category-level metric scale
  object shape and pose estimation,'' \emph{IEEE Robotics and Automation
  Letters}, vol.~6, no.~4, pp. 8575--8582, 2021.

\bibitem{chen2023stereopose}
K.~Chen, S.~James, C.~Sui, Y.-H. Liu, P.~Abbeel, and Q.~Dou, ``Stereopose:
  Category-level 6d transparent object pose estimation from stereo images via
  back-view nocs,'' in \emph{2023 IEEE International Conference on Robotics and
  Automation (ICRA)}.\hskip 1em plus 0.5em minus 0.4em\relax IEEE, 2023, pp.
  2855--2861.

\bibitem{sun2022onepose}
J.~Sun, Z.~Wang, S.~Zhang, X.~He, H.~Zhao, G.~Zhang, and X.~Zhou, ``Onepose:
  One-shot object pose estimation without cad models,'' in \emph{Proceedings of
  the IEEE/CVF Conference on Computer Vision and Pattern Recognition}, 2022,
  pp. 6825--6834.

\bibitem{he2022oneposeplusplus}
X.~He, J.~Sun, Y.~Wang, D.~Huang, H.~Bao, and X.~Zhou, ``Onepose++:
  Keypoint-free one-shot object pose estimation without {CAD} models,'' in
  \emph{Advances in Neural Information Processing Systems}, 2022.

\bibitem{EPnP}
V.~Lepetit, F.~Moreno-Noguer, and P.~Fua, ``Ep n p: An accurate o (n) solution
  to the p n p problem,'' \emph{International journal of computer vision},
  vol.~81, pp. 155--166, 2009.

\bibitem{nerf}
B.~Mildenhall, P.~P. Srinivasan, M.~Tancik, J.~T. Barron, R.~Ramamoorthi, and
  R.~Ng, ``Nerf: Representing scenes as neural radiance fields for view
  synthesis,'' in \emph{ECCV}, 2020.

\bibitem{inerf}
L.~Yen-Chen, P.~Florence, J.~T. Barron, A.~Rodriguez, P.~Isola, and T.-Y. Lin,
  ``inerf: Inverting neural radiance fields for pose estimation,'' in
  \emph{2021 IEEE/RSJ International Conference on Intelligent Robots and
  Systems (IROS)}.\hskip 1em plus 0.5em minus 0.4em\relax IEEE, 2021, pp.
  1323--1330.

\bibitem{lin2023parallel}
Y.~Lin, T.~M{\"u}ller, J.~Tremblay, B.~Wen, S.~Tyree, A.~Evans, P.~A. Vela, and
  S.~Birchfield, ``Parallel inversion of neural radiance fields for robust pose
  estimation,'' in \emph{2023 IEEE International Conference on Robotics and
  Automation (ICRA)}.\hskip 1em plus 0.5em minus 0.4em\relax IEEE, 2023, pp.
  9377--9384.

\bibitem{park2020latentfusion}
K.~Park, A.~Mousavian, Y.~Xiang, and D.~Fox, ``Latentfusion: End-to-end
  differentiable reconstruction and rendering for unseen object pose
  estimation,'' in \emph{Proceedings of the IEEE/CVF conference on computer
  vision and pattern recognition}, 2020, pp. 10\,710--10\,719.

\bibitem{palazzi2018end}
A.~Palazzi, L.~Bergamini, S.~Calderara, and R.~Cucchiara, ``End-to-end 6-dof
  object pose estimation through differentiable rasterization,'' in
  \emph{Proceedings of the European Conference on Computer Vision (ECCV)
  Workshops}, 2018, pp. 0--0.

\bibitem{chen2019learning}
W.~Chen, H.~Ling, J.~Gao, E.~Smith, J.~Lehtinen, A.~Jacobson, and S.~Fidler,
  ``Learning to predict 3d objects with an interpolation-based differentiable
  renderer,'' \emph{Advances in neural information processing systems},
  vol.~32, 2019.

\bibitem{yariv2021volume}
L.~Yariv, J.~Gu, Y.~Kasten, and Y.~Lipman, ``Volume rendering of neural
  implicit surfaces,'' \emph{Advances in Neural Information Processing
  Systems}, vol.~34, pp. 4805--4815, 2021.

\bibitem{yariv2020multiview}
L.~Yariv, Y.~Kasten, D.~Moran, M.~Galun, M.~Atzmon, B.~Ronen, and Y.~Lipman,
  ``Multiview neural surface reconstruction by disentangling geometry and
  appearance,'' \emph{Advances in Neural Information Processing Systems},
  vol.~33, pp. 2492--2502, 2020.

\bibitem{neus}
P.~Wang, L.~Liu, Y.~Liu, C.~Theobalt, T.~Komura, and W.~Wang, ``Neus: Learning
  neural implicit surfaces by volume rendering for multi-view reconstruction,''
  \emph{arXiv preprint arXiv:2106.10689}, 2021.

\bibitem{kehl2017ssd}
W.~Kehl, F.~Manhardt, F.~Tombari, S.~Ilic, and N.~Navab, ``Ssd-6d: Making
  rgb-based 3d detection and 6d pose estimation great again,'' in
  \emph{Proceedings of the IEEE international conference on computer vision},
  2017, pp. 1521--1529.

\bibitem{rad2017bb8}
M.~Rad and V.~Lepetit, ``Bb8: A scalable, accurate, robust to partial occlusion
  method for predicting the 3d poses of challenging objects without using
  depth,'' in \emph{Proceedings of the IEEE international conference on
  computer vision}, 2017, pp. 3828--3836.

\bibitem{tekin2018real}
B.~Tekin, S.~N. Sinha, and P.~Fua, ``Real-time seamless single shot 6d object
  pose prediction,'' in \emph{Proceedings of the IEEE conference on computer
  vision and pattern recognition}, 2018, pp. 292--301.

\bibitem{song2020hybridpose}
C.~Song, J.~Song, and Q.~Huang, ``Hybridpose: 6d object pose estimation under
  hybrid representations,'' in \emph{Proceedings of the IEEE/CVF conference on
  computer vision and pattern recognition}, 2020, pp. 431--440.

\bibitem{shugurov2022osop}
I.~Shugurov, F.~Li, B.~Busam, and S.~Ilic, ``Osop: A multi-stage one shot
  object pose estimation framework,'' in \emph{Proceedings of the IEEE/CVF
  Conference on Computer Vision and Pattern Recognition}, 2022, pp. 6835--6844.

\bibitem{megapose}
Y.~Labb\'e, L.~Manuelli, A.~Mousavian, S.~Tyree, S.~Birchfield, J.~Tremblay,
  J.~Carpentier, M.~Aubry, D.~Fox, and J.~Sivic, ``Megapose: 6d pose estimation
  of novel objects via render \& compare,'' in \emph{Proceedings of the 6th
  Conference on Robot Learning (CoRL)}, 2022.

\bibitem{ponimatkin2022focal}
G.~Ponimatkin, Y.~Labb{\'e}, B.~Russell, M.~Aubry, and J.~Sivic, ``Focal length
  and object pose estimation via render and compare,'' in \emph{Proceedings of
  the IEEE/CVF Conference on Computer Vision and Pattern Recognition}, 2022,
  pp. 3825--3834.

\bibitem{liu2019soft}
S.~Liu, T.~Li, W.~Chen, and H.~Li, ``Soft rasterizer: A differentiable renderer
  for image-based 3d reasoning,'' in \emph{Proceedings of the IEEE/CVF
  International Conference on Computer Vision}, 2019, pp. 7708--7717.

\bibitem{petersen2022gendr}
F.~Petersen, B.~Goldluecke, C.~Borgelt, and O.~Deussen, ``{GenDR: A Generalized
  Differentiable Renderer},'' in \emph{IEEE/CVF International Conference on
  Computer Vision and Pattern Recognition (CVPR)}, 2022.

\bibitem{maggio2023loc}
D.~Maggio, M.~Abate, J.~Shi, C.~Mario, and L.~Carlone, ``Loc-nerf: Monte carlo
  localization using neural radiance fields,'' in \emph{2023 IEEE International
  Conference on Robotics and Automation (ICRA)}.\hskip 1em plus 0.5em minus
  0.4em\relax IEEE, 2023, pp. 4018--4025.

\bibitem{sucar2021imap}
E.~Sucar, S.~Liu, J.~Ortiz, and A.~J. Davison, ``imap: Implicit mapping and
  positioning in real-time,'' in \emph{Proceedings of the IEEE/CVF
  International Conference on Computer Vision}, 2021, pp. 6229--6238.

\bibitem{zhu2022nice}
Z.~Zhu, S.~Peng, V.~Larsson, W.~Xu, H.~Bao, Z.~Cui, M.~R. Oswald, and
  M.~Pollefeys, ``Nice-slam: Neural implicit scalable encoding for slam,'' in
  \emph{Proceedings of the IEEE/CVF Conference on Computer Vision and Pattern
  Recognition}, 2022, pp. 12\,786--12\,796.

\bibitem{tosi2024nerfs}
F.~Tosi, Y.~Zhang, Z.~Gong, E.~Sandstr{\"o}m, S.~Mattoccia, M.~R. Oswald, and
  M.~Poggi, ``How nerfs and 3d gaussian splatting are reshaping slam: a
  survey,'' \emph{arXiv preprint arXiv:2402.13255}, 2024.

\bibitem{tang2012textured}
J.~Tang, S.~Miller, A.~Singh, and P.~Abbeel, ``A textured object recognition
  pipeline for color and depth image data,'' in \emph{2012 IEEE International
  Conference on Robotics and Automation}.\hskip 1em plus 0.5em minus
  0.4em\relax IEEE, 2012, pp. 3467--3474.

\bibitem{martinez2010moped}
M.~Martinez, A.~Collet, and S.~S. Srinivasa, ``Moped: A scalable and low
  latency object recognition and pose estimation system,'' in \emph{2010 IEEE
  International Conference on Robotics and Automation}.\hskip 1em plus 0.5em
  minus 0.4em\relax IEEE, 2010, pp. 2043--2049.

\bibitem{collet2009object}
A.~Collet, D.~Berenson, S.~S. Srinivasa, and D.~Ferguson, ``Object recognition
  and full pose registration from a single image for robotic manipulation,'' in
  \emph{2009 IEEE International Conference on Robotics and Automation}.\hskip
  1em plus 0.5em minus 0.4em\relax IEEE, 2009, pp. 48--55.

\bibitem{lowe2004distinctive}
D.~G. Lowe, ``Distinctive image features from scale-invariant keypoints,''
  \emph{International journal of computer vision}, vol.~60, pp. 91--110, 2004.

\bibitem{rosten2006machine}
E.~Rosten and T.~Drummond, ``Machine learning for high-speed corner
  detection,'' in \emph{Computer Vision--ECCV 2006: 9th European Conference on
  Computer Vision, Graz, Austria, May 7-13, 2006. Proceedings, Part I 9}.\hskip
  1em plus 0.5em minus 0.4em\relax Springer, 2006, pp. 430--443.

\bibitem{rublee2011orb}
E.~Rublee, V.~Rabaud, K.~Konolige, and G.~Bradski, ``Orb: An efficient
  alternative to sift or surf,'' in \emph{2011 International conference on
  computer vision}.\hskip 1em plus 0.5em minus 0.4em\relax Ieee, 2011, pp.
  2564--2571.

\bibitem{yi2018learning}
K.~M. Yi, E.~Trulls, Y.~Ono, V.~Lepetit, M.~Salzmann, and P.~Fua, ``Learning to
  find good correspondences,'' in \emph{Proceedings of the IEEE conference on
  computer vision and pattern recognition}, 2018, pp. 2666--2674.

\bibitem{ranftl2018deep}
R.~Ranftl and V.~Koltun, ``Deep fundamental matrix estimation,'' in
  \emph{Proceedings of the European conference on computer vision (ECCV)},
  2018, pp. 284--299.

\bibitem{sarlin2020superglue}
P.-E. Sarlin, D.~DeTone, T.~Malisiewicz, and A.~Rabinovich, ``Superglue:
  Learning feature matching with graph neural networks,'' in \emph{Proceedings
  of the IEEE/CVF conference on computer vision and pattern recognition}, 2020,
  pp. 4938--4947.

\bibitem{kajiya1984ray}
J.~T. Kajiya and B.~P. Von~Herzen, ``Ray tracing volume densities,'' \emph{ACM
  SIGGRAPH computer graphics}, vol.~18, no.~3, pp. 165--174, 1984.

\bibitem{loftr}
J.~Sun, Z.~Shen, Y.~Wang, H.~Bao, and X.~Zhou, ``Loftr: Detector-free local
  feature matching with transformers,'' in \emph{Proceedings of the IEEE/CVF
  conference on computer vision and pattern recognition}, 2021, pp. 8922--8931.

\bibitem{sarlin2021back}
P.-E. Sarlin, A.~Unagar, M.~Larsson, H.~Germain, C.~Toft, V.~Larsson,
  M.~Pollefeys, V.~Lepetit, L.~Hammarstrand, F.~Kahl \emph{et~al.}, ``Back to
  the feature: Learning robust camera localization from pixels to pose,'' in
  \emph{Proceedings of the IEEE/CVF conference on computer vision and pattern
  recognition}, 2021, pp. 3247--3257.

\bibitem{ransac}
M.~A. Fischler and R.~C. Bolles, ``Random sample consensus: a paradigm for
  model fitting with applications to image analysis and automated
  cartography,'' \emph{Communications of the ACM}, vol.~24, no.~6, pp.
  381--395, 1981.

\bibitem{llff}
B.~Mildenhall, P.~P. Srinivasan, R.~Ortiz-Cayon, N.~K. Kalantari,
  R.~Ramamoorthi, R.~Ng, and A.~Kar, ``Local light field fusion: Practical view
  synthesis with prescriptive sampling guidelines,'' \emph{ACM Transactions on
  Graphics (TOG)}, vol.~38, no.~4, pp. 1--14, 2019.

\bibitem{yugay2023gaussian}
V.~Yugay, Y.~Li, T.~Gevers, and M.~R. Oswald, ``Gaussian-slam: Photo-realistic
  dense slam with gaussian splatting,'' \emph{arXiv preprint arXiv:2312.10070},
  2023.

\bibitem{matsuki2023gaussian}
H.~Matsuki, R.~Murai, P.~H. Kelly, and A.~J. Davison, ``Gaussian splatting
  slam,'' \emph{arXiv preprint arXiv:2312.06741}, 2023.

\bibitem{yan2023gs}
C.~Yan, D.~Qu, D.~Wang, D.~Xu, Z.~Wang, B.~Zhao, and X.~Li, ``Gs-slam: Dense
  visual slam with 3d gaussian splatting,'' \emph{arXiv preprint
  arXiv:2311.11700}, 2023.

\bibitem{keetha2023splatam}
N.~Keetha, J.~Karhade, K.~M. Jatavallabhula, G.~Yang, S.~Scherer, D.~Ramanan,
  and J.~Luiten, ``Splatam: Splat, track \& map 3d gaussians for dense rgb-d
  slam,'' \emph{arXiv preprint arXiv:2312.02126}, 2023.

\end{thebibliography}

\end{document}